\documentclass[sn-basic]{sn-jnl}

\usepackage{graphicx}%
\usepackage{multirow}%
\usepackage{amsmath,amssymb,amsfonts}%
\usepackage{amsthm}%
\usepackage{mathrsfs}%
\usepackage[title]{appendix}%
\usepackage{xcolor}%
\usepackage{textcomp}%
\usepackage{manyfoot}%
\usepackage{booktabs}%
\usepackage{algorithm}%
\usepackage{algorithmicx}%
\usepackage{algpseudocode}%
\usepackage{listings}%

\theoremstyle{thmstyleone}%
\theoremstyle{thmstyletwo}%

\theoremstyle{thmstylethree}%

\begin{document}

\title[Structural-Temporal Coupling Anomaly Detection with Dynamic Graph Transformer]{Structural-Temporal Coupling Anomaly Detection with Dynamic Graph Transformer}


\author[1]{\fnm{Chang} \sur{Zong}}\email{zongchang@zust.edu.cn}

\author*[2]{\fnm{Yueting} \sur{Zhuang}}\email{yzhuang@zju.edu.cn}

\author[2]{\fnm{Jian} \sur{Shao}}\email{jshao@zju.edu.cn}

\author[2]{\fnm{Weiming} \sur{Lu}}\email{luwm@zju.edu.cn}

\affil[1]{\orgdiv{School of Computer Science}, \orgname{Zhejiang University of Science and Technology}, \orgaddress{ \city{Hangzhou}, \country{China}}}
\affil[2]{\orgdiv{School of Computer Science}, \orgname{Zhejiang University}, \orgaddress{ \city{Hangzhou}, \country{China}}}


\abstract{Detecting anomalous edges in dynamic graphs is an important task in many applications over evolving triple-based data, such as social networks, transaction management, and epidemiology. A major challenge with this task is the absence of structural-temporal coupling information, which decreases the ability of representation to distinguish anomalies from normal instances. Existing methods focus on handling independent structural and temporal features with embedding models, which ignore deep interaction between these two types of information. In this paper, we propose a structural-temporal coupling anomaly detection architecture with a dynamic graph transformer model. Specifically, we introduce structural and temporal features from two integration levels to provide anomaly-aware graph evolutionary patterns. Then, a dynamic graph transformer enhanced by two-dimensional positional encoding is implemented to capture both discrimination and contextual consistency signals. Extensive experiments on six datasets demonstrate that our method outperforms current state-of-the-art models. Finally, a case study illustrates the strength of our method when applied to a real-world task.}

\maketitle

\section{Introduction}
Dynamic graphs are ubiquitous and reflect the nature of temporal evolution in real-world scenarios. Detecting anomalous edges in dynamic graphs has become increasingly popular in recent years due to a variety of applications, including social networks, recommendation systems, and epidemiology \citep{skarding2021foundations}. Although various studies have been proposed on this task, anomaly detection in dynamic graphs is still a challenging task. One major challenge is the difficulty of using and learning discriminative knowledge from dynamic graphs where structural and temporal information are coupled \cite{liu2021anomaly}.

Previous works on solving this task can be summarized as feature-based methods and deep learning approaches. CM-Sketch \cite{ranshous2016scalable} combines the characteristics of the network structure with temporal indicators to detect abnormal edges. SpotLight \cite{eswaran2018spotlight} proposes a randomized sketching-based method to ensure that anomalous instances are mapped far from normal instances. These methods explicitly utilize structural information and temporal behavior, but their lack of representation limits detection performance. Deep learning methods have recently achieved better results. NetWalk \cite{yu2018netwalk} learns latent graph representations by using a number of network walks, followed by a clustering-based technique to detect network anomalies incrementally and dynamically. AddGraph \cite{zheng2019addgraph} and StrGNN \cite{cai2021structural} further exploit end-to-end deep neural network models to detect anomalies with the learning of structural and temporal information. These methods obey the architecture of learning graph embeddings followed by time-sequence modeling. TADDY \cite{liu2021anomaly}, the first method to apply a transformer model to anomaly detection in dynamic graphs, is capable of capturing informative representations from dynamic graphs with coupling structural-temporal patterns.

 Although considerable effort has been made to solve the task, there is still room for improvement in handling structural-temporal information. First, structural and temporal input features have not been well integrated, especially to describe anomalousness. CM-Sketch, StrGNN, and TADDY provide independent structure and time-series indicators, such as neighborhood scores, diffusion scores, and node distance, which cannot directly represent the behavior of anomalies in evolutionary graphs. Second, the architecture of concatenating graph embedding models with recurrent neural networks proposed in NetWalk, AddGraph, and StrGNN cannot utilize structural-temporal coupling information during training. The state-of-the-art approach, TADDY, employs a transformer to jointly encode structural and time-series embeddings of nodes. However, there is still a shortage of merging structural and temporal information in other processing steps, such as input features selection and positional encoding of transformer. All these questions on exploiting structural-temporal coupling information need to be studied further.

 To address the above challenges, we propose \textbf{STCAD}, a \underline{\textbf{S}}tructural-\underline{\textbf{T}}emporal \underline{\textbf{C}}oupling \underline{\textbf{A}}nomaly \underline{\textbf{D}}etection architecture using dynamic graph transformer. Specifically, STCAD presents a two-level feature encoding framework with three novel features of evolutionary graphs to capture anomaly-aware patterns in dynamic graphs. These features are discovered with an anomaly-targeted analysis strategy with benchmark datasets. Meanwhile, a two-dimensional positional encoding is proposed for the first time to label structural-temporal integrated positions in dynamic graphs. Absolute structural, relative structural, and temporal locations compose the position of a node in a dynamic graph. Finally, to utilize contextual consistency information in dynamic graphs, a mixed supervision training framework is implemented to capture anomaly discrimination and structural-temporal contextual consistency signals simultaneously. Our main contributions are summarized as follows.

\begin{itemize}
    \item A novel two-level anomaly-aware feature encoding framework is presented, which utilizes independent and coupling structural-temporal information to effectively detect anomalies in dynamic graphs. 
    \item A dynamic graph transformer model is presented to expressively learn node representations with structural-temporal integrated positions and is further applied to capture discrimination and consistency signals.
    \item Extensive experiments are conducted on six benchmark datasets and two different metrics. The results show that our proposed method outperforms existing state-of-the-art approaches.
\end{itemize}

\section{Related Work}
\subsection{Dynamic Graph Anomaly Detection}
Anomaly detection in dynamic graphs has become an active research field in recent years. The early method CM-Sketch \cite{ranshous2016scalable} describes an anomaly detection approach based on global and local structural properties, including a sample-based score, a preferential attachment score, and a homophily score to map each edge to its expected structural behavior. SpotLight \cite{eswaran2018spotlight} defines the distance between graphs in a sketch space to make anomalies far away from normal instances. The recent deep learning approach NetWalk \cite{yu2018netwalk} learns node embeddings from a deep autoencoder and a random walk strategy, followed by a clustering-based method to dynamically detect anomalies. AddGraph \cite{zheng2019addgraph} provides an end-to-end framework using a temporal graph convolutional network, which can capture long-term and short-term patterns in dynamic graphs. StrGNN \cite{cai2021structural} identifies the role of each node using the shortest path distance indicator. It generates embeddings from each snapshot with a graph convolution network and across timestamps with a gated recurrent unit technique. TADDY \cite{liu2021anomaly}, first introduces the transformer architecture to generate latent edge embeddings derived from structural and temporal properties of dynamic graphs. These methods are limited by a lack of guidance with structural-temporal coupling features, which we aim to study in this work.

\subsection{Transformer and Graph Transformer Models}
Transformer is a deep learning model that depends on the self-attention mechanism to learn representations from various input data. The transformer architecture is proposed originally for natural language data in \cite{vaswani2017attention}. In recent years, the transformer has been extended to handle other types of data. In the vision domain, transformer models are used to solve image classification \cite{dosovitskiy2020image,touvron2021training}, object detection \cite{carion2020end,zhu2020deformable}, and multimodal tasks \cite{li2020oscar,gui2021kat}. Graph transformers have been verified to perform well for graph representation learning. HGT \cite{hu2020heterogeneous} designs a method dependent on nodes and edge types to model heterogeneity over each edge. DySAT \cite{sankar2020dysat} captures dynamic structural evolution through joint self-attention with structural and temporal information. An absolute temporal location is used as the positional encoding. Graphit \cite{mialon2021graphit} and LSPE \cite{dwivedi2021graph} discuss the graph positional encoding based on the random walk diffusion process and graph Laplacian. Graphormer \cite{ying2021transformers} adds several effective structural encoding methods to a standard transformer model and shows superiority in graph representation tasks. The discussion of structural-temporal positional encoding in transformer models is absent in these methods, which should be explored to improve anomaly detection performance.

\section{Problem Formulation}

Following previous works \cite{liu2021anomaly} and \cite{cai2022temporal}, we define a graph that contains timestamps as $G=(V,E,T)$, where $V$ is the set of nodes, $E$ is the set of edges, $T$ is the number of timestamps. We denote a dynamic graph as a sequence of graph snapshots over $T$, denoted as $G^T=\lbrace S^1,S^2,...,S^T \rbrace$, where $S^t=(V^t,E^t, t)$ $(1 \le t \le T)$ is a snapshot with its nodes and edges at the timestamp $t$. An edge $e^t_{i,j}= (v^t_i,v^t_j) \in E^t$ indicates an edge between node $i$ and node $j$ at the timestamp $t$, where $v^t_i, v^t_j \in V^t$.

Given a dynamic graph composed of a set of sequential graph snapshots $G^T$ and a candidate edge $e^T_{i,j} \in E^T$, our goal is to learn a function $Score(\cdot)$ to calculate the anomaly score of the edge $e^T_{i,j}$ at time $T$ using its structural and historical information. A higher value of $Score(e^T_{i,j})$ means a higher probability that it is an anomalous edge.

\section{Methodology}
\begin{figure*}[!htb]
    \centering
    \includegraphics[width=13cm]{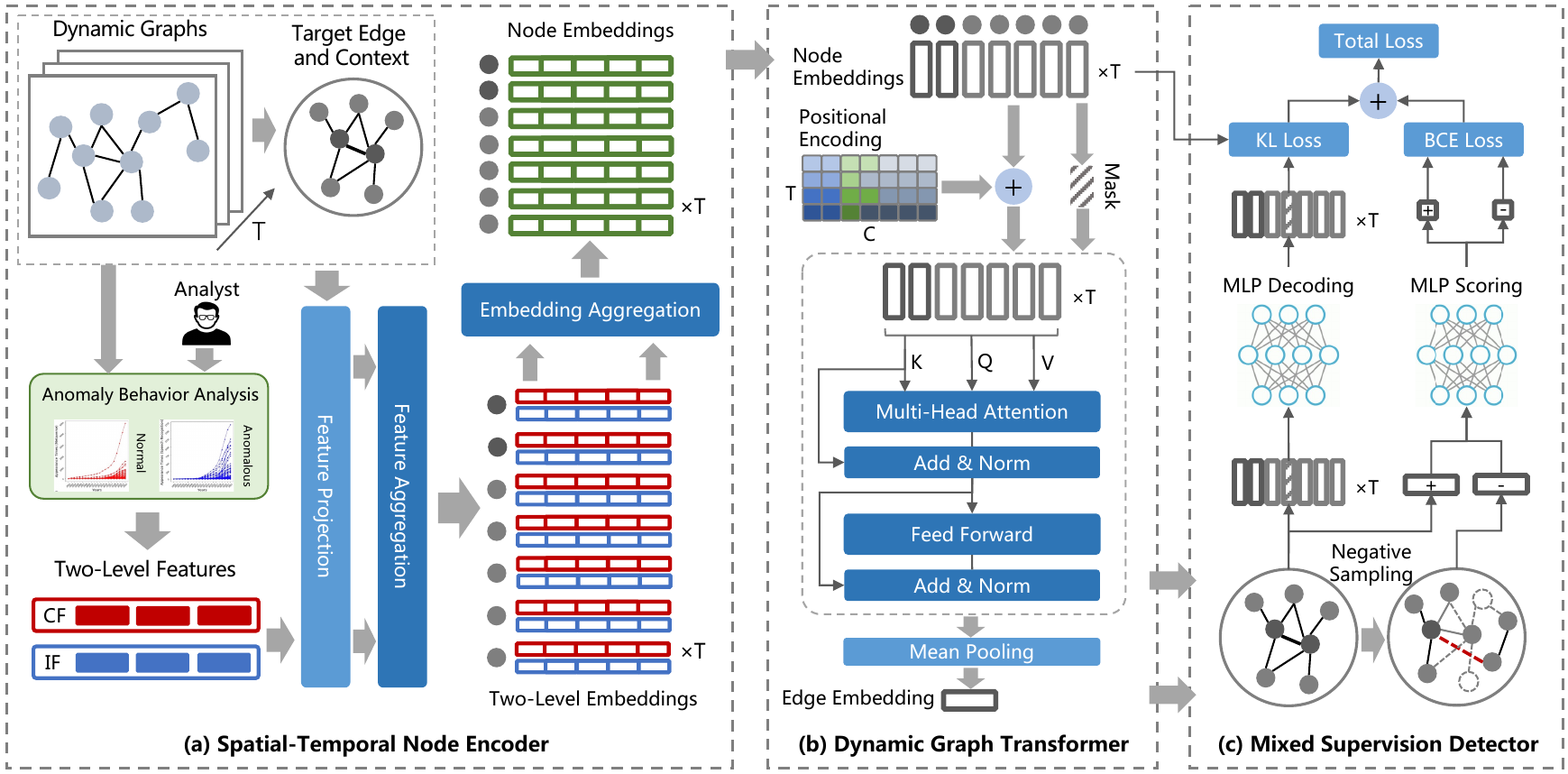}
    \caption{Architecture of our proposed STCAD for anomalous edges detection in dynamic graphs.}
    \label{architecture}
\end{figure*}

The motivation of our method is to further utilize structural-temporal coupling information from both input features and latent embeddings. The architecture of our method to detect anomalous edges in dynamic graphs is shown in Figure \ref{architecture}. Three modules are implemented, including a structural-temporal node encoder using two-level anomaly-aware features, a dynamic graph transformer with node sequences and two-dimensional positional encoding, and a mixed supervision detector with discriminative and contextual signals.

\subsection{Structural-Temporal Node Encoder}
In this module, we propose a structural-temporal node encoding process to obtain anomaly-aware input embeddings for the downstream model. 

\subsubsection{Edges and Context Sampling:} Candidate edges are randomly sampled from the last snapshot, with their one-hop context nodes as subgraphs along with timestamps. We denote $v^t_k \in \{\mathcal{N}(v^t_i), \mathcal{N}(v^t_j), v^t_i, v^t_j\}$ in the following paragraphs as the $k$-th node in the subgraph at time $t$ with the central edge $e^T_{i,j}$ , where $\mathcal{N}(v^t_i)$ and $\mathcal{N}(v^t_j)$ are the neighbors of the two nodes on the edge.

\subsubsection{Two-Level Feature Encoding:} To acquire structural-temporal knowledge to direct the following learning process, two-level features are selected. We first employ independent structural and temporal features to capture dynamic graph characteristics. The global and local structural features of a node can be computed with the PageRank and the shortest path algorithm, respectively. For temporal information, edge lifetime, which indicates the existence duration of an edge, is chosen for each node in a subgraph. Our first-level features are denoted as follows:
\begin{equation}
\begin{aligned}
\boldsymbol{F}_{glo}(v^t_k) &= PageRank(S^t, v^t_k) , \\
\boldsymbol{F}_{loc}(v^t_k) &= min(Dist(S^t, v^t_k, v^t_i), Dist(S^t, v^t_k, v^t_j)) , \\
\boldsymbol{F}_{tmp}(v^t_k) &= t - t_{start} ,\  t_{start} \leq t ,
\label{eq:eq1}
\end{aligned}
\end{equation} where $S^t$ is the snapshot at time $t$, $PageRank(\cdot)$ and $Dist(\cdot)$ are the PageRank and the shortest path distance functions, $t_{start}$ is the time when the edge $e_{i,j}^t$ firstly happens.

Our second-level features focus on structural-temporal coupling information. Different from features proposed in previous works, our anomaly-aware features are helpful for directly describing evolutionary anomaly behavior. We first discover that two nodes on an anomalous edge are usually further apart than normal ones before anomaly occurs. This distance change anomaly feature can be calculated as:

\begin{equation}
\begin{aligned}
\boldsymbol{F}_{dc}(v^t_k) = Dist(S^{t-\Delta t}, v^{t-\Delta t}_i, v^{t-\Delta t}_j) - 1 ,
\label{eq:eq3}
\end{aligned}
\end{equation} where $Dist(\cdot)$ is the function to compute the distance between $v^{t-\Delta t}_i$ and $v^{t-\Delta t}_j$ in the snapshot $S^{t-\Delta t}$ at the previous timestamp $t-\Delta t$, and the current distance is 1. Another indicator is that nodes from anomaly edges usually have smaller interaction changes among their surroundings than normal ones before anomaly links occur. This interaction change feature is denoted as:

\begin{equation}
\begin{aligned}
\boldsymbol{F}_{ic}&(v^t_k) =  | \left (Deg(S^t, v^t_i) + Deg(S^t, v^t_j)\right ) - \\
& \left (Deg(S^{t-\Delta t}, v^{t-\Delta t}_i) + Deg(S^{t-\Delta t}, v^{t-\Delta t}_j)\right ) |,
\label{eq:eq4}
\end{aligned}
\end{equation} where $Deg(\cdot)$ is the function for counting the number of degrees of each node from an edge. Moreover, we assume that nodes from anomalous edges generally have fewer common-neighbor changes than normal ones before the anomaly occurs, indicating evolutionary convergence in real-world dynamic graphs. This neighbor change feature is computed as follows.

\begin{equation}
\begin{aligned}
\boldsymbol{F}_{nc}(v^t_k) = | \sum\nolimits_{v \in \mathcal{N}^t_i \cap \mathcal{N}^t_j} 1 - \sum\nolimits_{v \in \mathcal{N}^{t-\Delta t}_i \cap \mathcal{N}^{t-\Delta t}_j} 1 |,
\label{eq:eq5}
\end{aligned}
\end{equation} where $v$ is the node from the intersection sets of $i$'s neighbors and $j$'s neighbors at two timestamps. To further explain, $\Delta t$ can be set to various values, and we find that 1 is an effective option in our work.

\subsubsection{Node Features Aggregation:} We project these two-level features into latent space with learnable parameters and aggregate them into a structural-temporal coupling embedding for each node, formulated as:

\begin{equation}
\begin{aligned}
\boldsymbol{X}_{level1}(v^t_k) = & \boldsymbol{F}_{glo}(v^t_k) \boldsymbol{W}_g + \boldsymbol{F}_{loc}(v^t_k) \boldsymbol{W}_l + \boldsymbol{F}_{tmp}(v^t_k) \boldsymbol{W}_t , \\
\boldsymbol{X}_{level2}(v^t_k) = & \boldsymbol{F}_{dc}(v^t_k) \boldsymbol{W}_d + \boldsymbol{F}_{ic}(v^t_k) \boldsymbol{W}_i + \boldsymbol{F}_{nc}(v^t_k) \boldsymbol{W}_c , \\
\boldsymbol{X}(v^t_k) &= \boldsymbol{X}_{level1}(v^t_k) + \boldsymbol{X}_{level2}(v^t_k) ,
\label{eq:eq6}
\end{aligned}
\end{equation} where $\boldsymbol{W}_g$, $\boldsymbol{W}_l$ and $\boldsymbol{W}_t \in \mathbb{R}^{1 \times d}$ are weights for the dynamic graph feature projection, $\boldsymbol{W}_d$, $\boldsymbol{W}_i$, and $\boldsymbol{W}_c \in \mathbb{R}^{1 \times d}$ are weights for projecting anomaly-aware evolutionary features.

\subsection{Dynamic Graph Transformer}
In this module, we propose a transformer model to generate edge embeddings with structural-temporal coupling features.

\subsubsection{Input Sequence Construction:} The input sequence of the transformer model is composed of contextual node embeddings of candidate edges. Specifically, for each timestamp, we generate a subsequence of node embeddings starting with two nodes from an edge followed by other context nodes. To ensure a fixed size of all the input sequences, we pad each subsequence with nodes from the central edge if there are fewer context nodes than the parameter we set. The input sequences are described below.

\begin{equation}
\begin{aligned}
Seq^T_C = \bigoplus_{t=1}^T[v^t_i, v^t_j, v^t_1, ..., v^t_C] ,
\label{eq:eq7}
\end{aligned}
\end{equation} where $\bigoplus$ is the concatenation operator, $C$ is the hyperparameter for the number of context nodes, $T$ is the number of timestamps.

\subsubsection{2D Positional Encoding:} Inspired by the positional encoding for handling image data in \cite{raisi20202d}, we propose a two-dimensional positional encoding for the first time to handle dynamic graphs. Similarly to the existing dynamic graph transformer model \cite{sankar2020dysat}, we employ an absolute temporal position as the first dimension of our positions. Furthermore, the structural position of a node can be represented by its location in a subgraph. We treat it as a piecewise function, which is a relative distance between the nodes in a subgraph and the nodes on the central edge. These two positions are denoted as follows.

\begin{equation}
\begin{aligned}
\boldsymbol{PE}_{tmp}(v^t_k) = t , \boldsymbol{PE}_{rel}(v^t_k) = \left\{ 
    \begin{array}{lc}
        0 &  v^t_k \in \{v^t_i, v^t_j\} , \\
        1 & v^t_k \in \mathcal{N}^t_i \cap \mathcal{N}^t_j , \\
        2 & others ,
    \end{array} \right.
\label{eq:eq8}
\end{aligned}
\end{equation} where $\boldsymbol{PE}_{tmp}(v^t_k)$ and $\boldsymbol{PE}_{rel}(v^t_k) \in \mathbb{R}^1$.

Then, each position is projected into a learnable embedding. Without using sinusoidal position encoding strategy in traditional methods, we directly map the positions in the temporal dimension and structural dimension into a space with a half size of hidden dimension as follows.

\begin{equation}
\begin{aligned}
\boldsymbol{PE}(v^t_k) &= \boldsymbol{PE}_{tmp}(v^t_k) \boldsymbol{W}_{tmp} \oplus \boldsymbol{PE}_{rel}(v^t_k) \boldsymbol{W}_{rel} ,
\label{eq:eq9}
\end{aligned}
\end{equation} where $\oplus$ is the binary concatenation operator, $\boldsymbol{W}_{tmp}$ and $\boldsymbol{W}_{rel} \in \mathbb{R}^{1 \times d/2}$ are weights for projecting the positions above into latent space.

\subsubsection{Transformer Edge Encoding:} With the node embedding $\boldsymbol{X}(v^t_k)$ we obtained in the previous section and our two-dimensional positional encoding $\boldsymbol{PE}(v^t_k)$, the input embedding of our transformer model is denoted as:

\begin{equation}
\begin{aligned}
\boldsymbol{H}(Seq^T_C) &= \bigoplus_{t=1}^T [\boldsymbol{In}(v^t_i), \boldsymbol{In}(v^t_j), \boldsymbol{In}(v^t_1), ..., \boldsymbol{In}(v^t_C)] , \\
\boldsymbol{In}(v^t_k) &= [\boldsymbol{X}(v^t_k)+\boldsymbol{PE}(v^t_k)]^\top, k \in \{i, j, 1, ..., C\}
\label{eq:eq10}
\end{aligned}
\end{equation} where $\boldsymbol{H}(Seq^T_C)$ is the input embedding of the sequence, $\boldsymbol{In}(v^t_k)$ is the input embedding of node $v^t_k$, and $[\cdot]^\top$ is the transpose operator. 

Our dynamic graph transformer encoder consists of a composition of transformer layers. Each layer contains a self-attention submodule and a position-wise feedforward network. The input embedding sequence is projected by three matrices to form the corresponding representations. The attention $\boldsymbol{A}^{(l)}$ from the $l$-th layer of the self-attention module is formulated as follows.

\begin{equation}
\begin{aligned}
\boldsymbol{A}^{(l)} = Softmax(\frac{\boldsymbol{Q}^{(l)} \boldsymbol{K}^{(l)\top}}{\sqrt{d'}}) \boldsymbol{V}^{(l)} , \boldsymbol{Q}^{(l)} = \boldsymbol{H}(Seq_C^T)^{(l-1)} \boldsymbol{W}_Q^{(l)} , \\
\boldsymbol{K}^{(l)} = \boldsymbol{H}(Seq^T_C)^{(l-1)} \boldsymbol{W}_K^{(l)} , \boldsymbol{V}^{(l)} = \boldsymbol{H}(Seq^T_C)^{(l-1)} \boldsymbol{W}_V^{(l)} ,
\label{eq:eq11}
\end{aligned}
\end{equation} where $\boldsymbol{W}_Q^{(l)} \in \mathbb{R}^{d \times d'}$, $\boldsymbol{W}_K^{(l)} \in \mathbb{R}^{d \times d'}$, $\boldsymbol{W}_V^{(l)} \in \mathbb{R}^{d \times d'}$, $d'$ is the number of attention heads times the hidden dimension, $\boldsymbol{H}(Seq_C^T)^{(l-1)}$ is the sequence of embeddings updated from the previous transformer layer, and $\boldsymbol{H}(Seq_C^T)^{(0)} = \boldsymbol{H}(Seq^T_C)$

A feed-forward network is applied to our normalized attention and residual, followed by another normalization layer, as described below.

\begin{equation}
\begin{aligned}
\boldsymbol{H}^{(l)} = LN(\boldsymbol{A}^{(l)} + \boldsymbol{Q}^{(l)}) , \boldsymbol{H}_{out}^{(l)} = LN(FFN(\boldsymbol{H}^{(l)}) + \boldsymbol{H}^{(l)}) ,
\label{eq:eq12}
\end{aligned}
\end{equation} where $LN(\cdot)$ is the layer normalization method, $FFN(\cdot)$ is the feed-forward network block.

Finally, we obtain the edge embedding by taking the average value of all the node embeddings from the entire sequence as:

\begin{equation}
\begin{aligned}
\boldsymbol{H}(e^T_{i,j}) = \frac{1}{K} \sum_{k=1}^{K} \left(\boldsymbol{H}_{out}^{(L)}\right)_k ,
\label{eq:eq13}
\end{aligned}
\end{equation} where $\left(\boldsymbol{H}_{out}^{(L)}\right)_k$ is the $k$-th embedding from the output of $L$-th transformer layer, $K=(C+2) \times T$ is the total length of a node sequence.

\subsection{Mixed Supervision Anomaly Detector}
In this module, we compute the anomaly scores of the candidate edges and train the model in a mixed supervision mode to capture the signals of discrimination and contextual consistency.

\subsubsection{Discriminative Anomaly Detector:} In order to implement an end-to-end anomaly detection framework, a common strategy is to calculate the anomaly score for each edge and train the model discriminatively. As there is no ground-truth anomalous sample in our datasets, we use the anomaly injection strategy introduced in \cite{yu2018netwalk}. A straightforward negative sampling method is implemented to select edges that have never occurred during all timestamps as anomalies. All existing edges in our current snapshot are viewed as normal samples (labeled 0), and the same number of anomalous edges (labeled 1) are sampled.

With the labeling strategy above, we should limit edge embeddings into a single score between 0 and 1 to fit labels. The anomaly score of an edge is calculated as below.

\begin{equation}
\begin{aligned}
Score(e^T_{i,j}) = Sigmoid(Linear(\boldsymbol{H}(e^T_{i,j})) ,
\label{eq:eq14}
\end{aligned}
\end{equation} where $Linear(\cdot)$ is the Linear function to project edge embeddings into a single value, $Sigmoid(\cdot)$ is the Sigmoid function to limit the range of the value.

A binary cross-entropy loss is applied to quantify the distance between anomaly scores and labels. Our discriminative loss is defined as:

\begin{equation}
\begin{aligned}
\mathcal{L}_{dis} = -\sum_{n=1}^N log(1-Score_{i,j}^+)+log(Score_{i,j}^-) ,
\label{eq:eq15}
\end{aligned}
\end{equation} where $N$ is the total number of samples, $Score_{i,j}^+$ and $Score_{i,j}^-$ are anomaly scores of positive and negative samples, respectively.

\subsubsection{Contextual Anomaly Detector:} Meanwhile, illuminated by \cite{Zhang2022ReconstructionEM} which utilize the consistency signal in attributed graphs, we assume that anomaly instances usually present inconsistency in terms of structural and temporal context in dynamic graphs. This additional signal can be converted to our second anomaly detector by reconstructing the entire input sequence with an encoder-decoder framework. Specifically, we randomly mask one of the elements from the input sequence by replacing it with an all-zero embedding instead. After being encoded with our transformer layers, a multilayer perceptron (MLP) model is employed as a decoder to try to recover the embedding sequence from the masked one. A pointwise KL-divergence function is then used to calculate the contextual loss as follows.

\begin{equation}
\begin{aligned}
\mathcal{L}_{con} = \sum_{n=1}^{N} \boldsymbol{H}(Seq^T_C) log \frac{\boldsymbol{H}(Seq^T_C)}{Dec(Enc(\mathcal{M}(\boldsymbol{H}(Seq^T_C))))} ,
\label{eq:eq16}
\end{aligned}
\end{equation} where $\boldsymbol{H}(Seq^T_C)$, introduced in Section 4.2, is the original embedding of our input sequence, $\mathcal{M}(\cdot)$ is a mask function, $Enc(\cdot)$ is our transformer encoder and $Dec(\cdot)$ is our MLP decoder, $N$ is the total number of samples.

\subsubsection{Total Training Loss:} Our final loss function is a combination of discrimination loss and contextual consistency loss. We sum the two losses as follows:

\begin{equation}
\begin{aligned}
\mathcal{L} = \mathcal{L}_{dis} + \lambda\mathcal{L}_{con} ,
\label{eq:eq17}
\end{aligned}
\end{equation} where $\lambda$ is a hyperparameter to balance these two losses.

\subsection{Overall Training Algorithm}
We train our method in an iterative end-to-end manner. Features and samples are generated prior to our training process. First, embeddings of nodes from subgraphs and time sequences are calculated. Then, the hidden embeddings of the node sequence are learned with our transformer model. Finally, the mixed supervision loss is computed for training. Model parameters are updated by backpropagation. The overall training procedure is described in Algorithm \ref{algorithm}.

\begin{algorithm}
    \caption{STCAD training algorithm}
    \label{algorithm}
    \textbf{Input:} Dynamic Graph: $G^T=\lbrace S^1,S^2,...,S^T \rbrace$ \\
    \textbf{Parameters:} \\
    Embedding Dimension: $D$, Context Number: $C$, Time Sequence Length: $T$, Epoch Number: $Epoch$, Loss Balancer: $\lambda$ \\
    \textbf{Output:} Training Loss: $\mathcal{L}$, Anomaly Score: $Score$
    \begin{algorithmic}[1] 
    \State Sampling nodes $v^t_k \in \{v^t_i, v^t_j, \mathcal{N}(v^t_i), \mathcal{N}(v^t_j)\}$ from positive and negative edges
    \State Calculate independent features $\boldsymbol{F}_{glo}(v^t_k)$,  $\boldsymbol{F}_{loc}(v^t_k)$, $\boldsymbol{F}_{tmp}(v^t_k)$ with Equation (1)
    \State Calculate structural-temporal coupling features $\boldsymbol{F}_{dc}(v^t_k)$, $\boldsymbol{F}_{ic}(v^t_k)$, $\boldsymbol{F}_{cn}(v^t_k)$ with Equation (2), (3) and (4)
    \For{each $Epoch$}
        \For{each $e^T_{i,j}$}
            \State Aggregate node embeddings $\boldsymbol{X}(v^t_k)$ with Equation (5)
            \State Construct input sequence $Seq^T_C$ with Equation (6)
            \State Calculate 2D positional embedding $\boldsymbol{PE}(v^t_k)$ with Equation (7) and (8)
            \State Generate transformer input embedding $\boldsymbol{H}(Seq^T_C)$ with Equation (9)
            \State Calculate transformer output embedding $\boldsymbol{H}_{out}^{(l)}$ with Equation (10) and (11)
            \State Obtain edge embedding $\boldsymbol{H}(e^T_{i,j})$ with Equation (12)
            \State Calculate anomaly score $Score(e^T_{i,j})$ with Equation (13)
            \State Calculate mixed training loss $\mathcal{L}$ with Equation (14), (15) and (16)
            \State Return $\mathcal{L}$, $Score(e^T_{i,j})$
        \EndFor
        \State Optimize parameters with back propagation and Adam
    \EndFor
    \end{algorithmic}
\end{algorithm}

\section{Experiments}

\subsection{Benchmark Datasets}
Six benchmark dynamic graph datasets are used for our evaluation, including a message interaction dataset \textbf{UCI Message} \cite{opsahl2009clustering}, a news comment interaction dataset \textbf{Digg} \cite{de2009social}, an email transmission dataset \textbf{Email-DNC} \cite{rossi2015network}, an autonomous system connection dataset \textbf{AS-Topology} \cite{zhang2005collecting}, and two Bitcoin platform rating datasets \textbf{Bitcoin-Alpha} \cite{kumar2016edge} and \textbf{Bitcoin-OTC} \cite{kumar2018rev2}. Statistical information is shown in Table \ref{dataset}. 

\begin{table}[!htb]
    \centering
    \caption{The number of nodes and edges in each benchmark dataset. Edges which have the same nodes but with different timestamps are regarded as different instances.}
    \label{dataset}
    \begin{tabular}{lll}
        \hline
        \textbf{Dataset}  & \textbf{Number of Nodes} & \textbf{Number of Edges} \\
        \hline
        UCI-Message     &   1899   &  59798\\
        Digg     &   30398   &   87626\\
        Email-DNC  &   1891   &  32880\\
        Bitcoin-Alpha &   3783   &  24186\\
        Bitcoin-OTC &   5881   &  35592\\
        AS-Topology &   37461   &  155507\\
        \hline
    \end{tabular}
\end{table}

\subsection{Experimental Settings}
\subsubsection{Baselines:} We select three types of approach as our baseline methods for comparing performance. For feature-based methods, we first implement a simple method that uses a combination of structural and temporal properties with linear layers named \textbf{STFeature}. We further replace these properties with the features proposed in CM-Sketch \cite{ranshous2016scalable}, denoted as \textbf{CMFeature}. For separated embedding methods, we test the performance discussed in the work \textbf{AddGraph} \cite{zheng2019addgraph} and \textbf{StrGNN} \cite{cai2021structural}. For dynamic graph transformer method, we adopt \textbf{TADDY} \cite{liu2021anomaly}, the state-of-the-art method, as our competitor.

\subsubsection{Implementation Details:} Each dataset is divided into successive snapshots along with timestamps. We use the first half of the sequences as our training set and the second half for testing. All existing edges in the last snapshot are labeled as normal (negative), and we sample the same number of edges that never occur from previous snapshots as anomalies (positive). In the testing set, we follow \cite{yu2018netwalk} to perform anomaly injections with 1\%, 5\% and 10\% proportional rates. We set the time sequence length to 4 and the number of subgraph context nodes to 5. The dimensions of projected and hidden embeddings are set to 32. The loss balancer is set to 1.0. The numbers of attention layer and heads are set to 2. We apply the Adam optimizer with a learning rate of 0.001. We use ROC-AUC and AP as our evaluation metrics to consider both precision and recall, as anomalous targets are usually much fewer than normal ones. We train our model on an NVIDIA RTX 3090 Ti GPU and select the best result from 300 training epochs for each experiment. The size of each snapshot is 4000 for UCI-Message, 6000 for Digg, 3000 for Email-DNC, 2000 for Bitcoin-Alpha and Bitcoin-OTC, and 8000 for AS-Topology.

\subsection{Results and Analysis}
Table \ref{performance} shows the results of the performance comparison on six benchmark datasets in terms of AUC and AP values. As shown in the table, our proposed STCAD outperforms all baselines on six datasets with both metrics with an average increase of 3.21\% on AUC and 53.67\% on AP. This significant difference demonstrates that STCAD is much better at capturing fewer positive samples due to our structural-temporal coupling method. Deep learning models generally perform better than the feature-based model in all experiments. However, TADDY with the ordinary transformer model cannot always surpass other deep learning methods on some datasets (Bitcoin-Alpha and Bitcoin-OTC), which indicates that the exploitation of structural-temporal information is more important in this task. Meanwhile, our STCAD has a consistent performance gain for all proportions of anomalies injection. A slightly larger performance gain occurs in 1\% injection proportion than 5\% and 10\%, which shows the effectiveness of our model in handling the sparsity of anomaly.

\begin{sidewaystable}
\caption{Performance reported on six benchmark datasets with AUC and AP metrics. Three anomaly proportions are evaluated. The best result for each experiment is highlighted in bold.}
\label{performance}
\begin{tabular}{|l|lll|lll|lll|lll|lll|lll|}
\hline
\multirow{3}{*}{\textbf{Methods}}  & \multicolumn{6}{c|}{\textbf{UCI-Message}}  &\multicolumn{6}{c|}{\textbf{Digg}}
\\  & 
\multicolumn{3}{c|}{\textbf{AUC}} & \multicolumn{3}{c|}{\textbf{AP}} & \multicolumn{3}{c|}{\textbf{AUC}}  &\multicolumn{3}{c|}{\textbf{AP}}
\\  & 
\multicolumn{1}{l|}{\textbf{1\%}} & \multicolumn{1}{l|}{\textbf{5\%}} & \textbf{10\%} & \multicolumn{1}{l|}{\textbf{1\%}} & \multicolumn{1}{l|}{\textbf{5\%}} & \textbf{10\%} & \multicolumn{1}{l|}{\textbf{1\%}} & \multicolumn{1}{l|}{\textbf{5\%}} & \textbf{10\%} & \multicolumn{1}{l|}{\textbf{1\%}} & \multicolumn{1}{l|}{\textbf{5\%}} & \textbf{10\%} 
\\ \hline      
\multirow{1}{*}{STFeature} & \multicolumn{1}{l|}{0.8514} & \multicolumn{1}{l|}{0.8469} & 0.8595 & \multicolumn{1}{l|}{0.2855} & \multicolumn{1}{l|}{0.4843} & 0.6004 & \multicolumn{1}{l|}{0.8144} & \multicolumn{1}{l|}{0.8350} & 0.8459 & \multicolumn{1}{l|}{0.1955} & \multicolumn{1}{l|}{0.4002} & 0.5043
\\
\multirow{1}{*}{CMFeature} & \multicolumn{1}{l|}{0.8678} & \multicolumn{1}{l|}{0.8508} & 0.8522 & \multicolumn{1}{l|}{0.3252} & \multicolumn{1}{l|}{0.5296} & 0.6354 & \multicolumn{1}{l|}{0.8078} & \multicolumn{1}{l|}{0.8190} & 0.8235 & \multicolumn{1}{l|}{0.2124} & \multicolumn{1}{l|}{0.4285} & 0.5310
\\
\multirow{1}{*}{AddGraph} & \multicolumn{1}{l|}{0.8916} & \multicolumn{1}{l|}{0.9064} & 0.9042 & \multicolumn{1}{l|}{0.3977} & \multicolumn{1}{l|}{0.6238} & 0.7369 & \multicolumn{1}{l|}{0.9182} & \multicolumn{1}{l|}{0.9311} & 0.9104 & \multicolumn{1}{l|}{0.3662} & \multicolumn{1}{l|}{0.5308} & 0.6135       
\\  
\multirow{1}{*}{StrGNN} & \multicolumn{1}{l|}{0.9080} & \multicolumn{1}{l|}{0.9226} & 0.9313 & \multicolumn{1}{l|}{0.4231} & \multicolumn{1}{l|}{0.6790} & 0.7669 & \multicolumn{1}{l|}{0.9098} & \multicolumn{1}{l|}{0.9224} & 0.9223 & \multicolumn{1}{l|}{0.3541} & \multicolumn{1}{l|}{0.5279} & 0.6636
\\
\multirow{1}{*}{TADDY} & \multicolumn{1}{l|}{0.9620} & \multicolumn{1}{l|}{0.9437} & 0.9496 & \multicolumn{1}{l|}{0.4854} & \multicolumn{1}{l|}{0.6665} & 0.7697 & \multicolumn{1}{l|}{0.9026} & \multicolumn{1}{l|}{0.9139} & 0.9175 & \multicolumn{1}{l|}{0.3045} & \multicolumn{1}{l|}{0.5083} & 0.6102
\\ \hline  
\multirow{1}{*}{\textbf{STCAD}} & \multicolumn{1}{l|}{\textbf{0.9896}} & \multicolumn{1}{l|}{\textbf{0.9715}} & \textbf{0.9735} & \multicolumn{1}{l|}{\textbf{0.6748}} & \multicolumn{1}{l|}{\textbf{0.8027}} & \textbf{0.8764} & \multicolumn{1}{l|}{\textbf{0.9709}} & \multicolumn{1}{l|}{\textbf{0.9828}} & \textbf{0.9603} & \multicolumn{1}{l|}{\textbf{0.7335}} & \multicolumn{1}{l|}{\textbf{0.8685}} & \textbf{0.8487}
\\ \hline 
\end{tabular}

\begin{tabular}{|l|lll|lll|lll|lll|lll|lll|}
\hline
\multirow{3}{*}{\textbf{Methods}}  & \multicolumn{6}{c|}{\textbf{Email-DNC}}  &\multicolumn{6}{c|}{\textbf{AS-Topology}}
\\  & 
\multicolumn{3}{c|}{\textbf{AUC}} & \multicolumn{3}{c|}{\textbf{AP}} & \multicolumn{3}{c|}{\textbf{AUC}}  &\multicolumn{3}{c|}{\textbf{AP}}
\\  & 
\multicolumn{1}{l|}{\textbf{1\%}} & \multicolumn{1}{l|}{\textbf{5\%}} & \textbf{10\%} & \multicolumn{1}{l|}{\textbf{1\%}} & \multicolumn{1}{l|}{\textbf{5\%}} & \textbf{10\%} & \multicolumn{1}{l|}{\textbf{1\%}} & \multicolumn{1}{l|}{\textbf{5\%}} & \textbf{10\%} & \multicolumn{1}{l|}{\textbf{1\%}} & \multicolumn{1}{l|}{\textbf{5\%}} & \textbf{10\%} 
\\ \hline      
\multirow{1}{*}{STFeature} & \multicolumn{1}{l|}{0.9141} & \multicolumn{1}{l|}{0.9030} & 0.8849 & \multicolumn{1}{l|}{0.0645} & \multicolumn{1}{l|}{0.2577} & 0.3919 & \multicolumn{1}{l|}{0.9400} & \multicolumn{1}{l|}{0.9296} & 0.9346 & \multicolumn{1}{l|}{0.4215} & \multicolumn{1}{l|}{0.5345} & 0.7004
\\
\multirow{1}{*}{CMFeature} & \multicolumn{1}{l|}{0.9219} & \multicolumn{1}{l|}{0.9141} & 0.8940 & \multicolumn{1}{l|}{0.0732} & \multicolumn{1}{l|}{0.2405} & 0.4144 & \multicolumn{1}{l|}{0.9577} & \multicolumn{1}{l|}{0.9452} & 0.9479 & \multicolumn{1}{l|}{0.4938} & \multicolumn{1}{l|}{0.6120} & 0.7583
\\
\multirow{1}{*}{AddGraph} & \multicolumn{1}{l|}{0.9282} & \multicolumn{1}{l|}{0.9174} & 0.9107 & \multicolumn{1}{l|}{0.0695} & \multicolumn{1}{l|}{0.2609} & 0.4041 & \multicolumn{1}{l|}{0.9574} & \multicolumn{1}{l|}{0.9549} & 0.9546 & \multicolumn{1}{l|}{0.5877} & \multicolumn{1}{l|}{0.6706} & 0.7701       
\\  
\multirow{1}{*}{StrGNN} & \multicolumn{1}{l|}{0.9353} & \multicolumn{1}{l|}{0.9291} & 0.9234 & \multicolumn{1}{l|}{0.0707} & \multicolumn{1}{l|}{0.2655} & 0.4087 & \multicolumn{1}{l|}{0.9693} & \multicolumn{1}{l|}{0.9658} & 0.9657 & \multicolumn{1}{l|}{0.6549} & \multicolumn{1}{l|}{0.7449} & 0.8239
\\
\multirow{1}{*}{TADDY} & \multicolumn{1}{l|}{0.9454} & \multicolumn{1}{l|}{0.9548} & 0.9543 & \multicolumn{1}{l|}{0.1038} & \multicolumn{1}{l|}{0.3998} & 0.5742 & \multicolumn{1}{l|}{0.9758} & \multicolumn{1}{l|}{0.9777} & 0.9807 & \multicolumn{1}{l|}{0.7541} & \multicolumn{1}{l|}{0.8297} & 0.8955
\\ \hline  
\multirow{1}{*}{\textbf{STCAD}} & \multicolumn{1}{l|}{\textbf{0.9920}} & \multicolumn{1}{l|}{\textbf{0.9953}} & \textbf{0.9897} & \multicolumn{1}{l|}{\textbf{0.6061}} & \multicolumn{1}{l|}{\textbf{0.9140}} & \textbf{0.9231} & \multicolumn{1}{l|}{\textbf{0.9836}} & \multicolumn{1}{l|}{\textbf{0.9864}} & \textbf{0.9890} & \multicolumn{1}{l|}{\textbf{0.8163}} & \multicolumn{1}{l|}{\textbf{0.8752}} & \textbf{0.9329}
\\ \hline 
\end{tabular}

\begin{tabular}{|l|lll|lll|lll|lll|lll|lll|}
\hline
\multirow{3}{*}{\textbf{Methods}}  & \multicolumn{6}{c|}{\textbf{Bitcoin-Alpha}}  &\multicolumn{6}{c|}{\textbf{Bitcoin-OTC}}
\\  & 
\multicolumn{3}{c|}{\textbf{AUC}} & \multicolumn{3}{c|}{\textbf{AP}} & \multicolumn{3}{c|}{\textbf{AUC}}  &\multicolumn{3}{c|}{\textbf{AP}}
\\  & 
\multicolumn{1}{l|}{\textbf{1\%}} & \multicolumn{1}{l|}{\textbf{5\%}} & \textbf{10\%} & \multicolumn{1}{l|}{\textbf{1\%}} & \multicolumn{1}{l|}{\textbf{5\%}} & \textbf{10\%} & \multicolumn{1}{l|}{\textbf{1\%}} & \multicolumn{1}{l|}{\textbf{5\%}} & \textbf{10\%} & \multicolumn{1}{l|}{\textbf{1\%}} & \multicolumn{1}{l|}{\textbf{5\%}} & \textbf{10\%} 
\\ \hline      
\multirow{1}{*}{STFeature} & \multicolumn{1}{l|}{0.9110} & \multicolumn{1}{l|}{0.9294} & 0.9409 & \multicolumn{1}{l|}{0.3723} & \multicolumn{1}{l|}{0.6679} & 0.7710 & \multicolumn{1}{l|}{0.9588} & \multicolumn{1}{l|}{0.9468} & 0.9465 & \multicolumn{1}{l|}{0.6551} & \multicolumn{1}{l|}{0.7015} & 0.7970
\\
\multirow{1}{*}{CMFeature} & \multicolumn{1}{l|}{0.9004} & \multicolumn{1}{l|}{0.9155} & 0.9279 & \multicolumn{1}{l|}{0.4248} & \multicolumn{1}{l|}{0.6720} & 0.7641 & \multicolumn{1}{l|}{0.9433} & \multicolumn{1}{l|}{0.9309} & 0.9302 & \multicolumn{1}{l|}{0.6378} & \multicolumn{1}{l|}{0.6749} & 0.7739
\\
\multirow{1}{*}{AddGraph} & \multicolumn{1}{l|}{0.9140} & \multicolumn{1}{l|}{0.9322} & 0.9546 & \multicolumn{1}{l|}{0.4017} & \multicolumn{1}{l|}{0.7280} & 0.8193 & \multicolumn{1}{l|}{0.9825} & \multicolumn{1}{l|}{0.9736} & 0.9629 & \multicolumn{1}{l|}{0.8033} & \multicolumn{1}{l|}{0.8898} & 0.8917 
\\  
\multirow{1}{*}{StrGNN} & \multicolumn{1}{l|}{0.9148} & \multicolumn{1}{l|}{0.9476} & 0.9618 & \multicolumn{1}{l|}{0.6166} & \multicolumn{1}{l|}{0.8218} & 0.8982 & \multicolumn{1}{l|}{0.9809} & \multicolumn{1}{l|}{0.9704} & 0.9701 & \multicolumn{1}{l|}{0.7868} & \multicolumn{1}{l|}{0.8766} & 0.9152
\\
\multirow{1}{*}{TADDY} & \multicolumn{1}{l|}{0.9355} & \multicolumn{1}{l|}{0.9647} & 0.9732 & \multicolumn{1}{l|}{0.4298} & \multicolumn{1}{l|}{0.7663} & 0.8573 & \multicolumn{1}{l|}{0.9797} & \multicolumn{1}{l|}{0.9608} & 0.9628 & \multicolumn{1}{l|}{0.7041} & \multicolumn{1}{l|}{0.7769} & 0.8595
\\ \hline  
\multirow{1}{*}{\textbf{STCAD}} & \multicolumn{1}{l|}{\textbf{0.9711}} & \multicolumn{1}{l|}{\textbf{0.9863}} & \textbf{0.9922} & \multicolumn{1}{l|}{\textbf{0.8036}} & \multicolumn{1}{l|}{\textbf{0.9404}} & \textbf{0.9700} & \multicolumn{1}{l|}{\textbf{0.9919}} & \multicolumn{1}{l|}{\textbf{0.9897}} & \textbf{0.9861} & \multicolumn{1}{l|}{\textbf{0.8862}} & \multicolumn{1}{l|}{\textbf{0.9206}} & \textbf{0.9501}
\\ \hline 
\end{tabular}
\end{sidewaystable}

\subsection{Ablation Study}
We further conduct an ablation study to prove the effectiveness of each component in STCAD. We denote STCAD-IF as the method without structural and temporal independent features and STCAD-CF as the method without structural-temporal coupling features. STCAD-PE is defined as the method for using the transformer without our 2D positional encoding. STCAD-SPE and STCAD-TPE are models with elimination of structural and temporal positions, respectively. The method STCAD-SSL is the model without self-supervised loss to capture contextual consistency. We select four datasets from different domains with 10\% anomaly injection for analysis. The performance is described in Table \ref{ablation}. The results show that all components contribute to the task consistently. Our proposed structural-temporal coupling features outperform independent features with an average of 1.38\% on AUC and 5.41\% on AP. Our 2D positional encoding is a crucial module, and the temporal position generally plays a greater role than the structural position. Meanwhile, capturing contextual consistency signals is beneficial for the overall performance of STCAD.

\begin{table}[!htb]
\centering
\caption{The AUC and AP values of ablation study with three types of variants of STCAD on four representative datasets. A 10\% anomaly injection is applied.}
\label{ablation}
\begin{tabular}{|l|ll|ll|ll|ll|}
\hline
\multirow{2}{*}{\textbf{Variants}} &\multicolumn{2}{c|}{\textbf{UCI-Message}} &\multicolumn{2}{c|}{\textbf{Digg}}  &\multicolumn{2}{c|}{\textbf{AS-Topology}}  &\multicolumn{2}{c|}{\textbf{Bitcoin-Alpha}} 
\\  & 
\multicolumn{1}{c|}{\textbf{AUC}} & \multicolumn{1}{c|}{\textbf{AP}} & \multicolumn{1}{c|}{\textbf{AUC}} & \multicolumn{1}{c|}{\textbf{AP}} & \multicolumn{1}{c|}{\textbf{AUC}}  &\multicolumn{1}{c|}{\textbf{AP}}   &\multicolumn{1}{c|}{\textbf{AUC}} &\multicolumn{1}{c|}{\textbf{AP}}
\\ \hline      
\multirow{1}{*}{STCAD-IF} & \multicolumn{1}{l|}{0.9667} & \multicolumn{1}{l|}{0.8593} & \multicolumn{1}{l|}{0.9546} & \multicolumn{1}{l|}{0.8417} & \multicolumn{1}{l|}{0.9845} & \multicolumn{1}{l|}{0.9241} & \multicolumn{1}{l|}{0.9839} & \multicolumn{1}{l|}{0.9424}
\\
\multirow{1}{*}{STCAD-CF} & \multicolumn{1}{l|}{0.9598} & \multicolumn{1}{l|}{0.7625} & \multicolumn{1}{l|}{0.9559} & \multicolumn{1}{l|}{0.7202} & \multicolumn{1}{l|}{0.9820} & \multicolumn{1}{l|}{0.9122} & \multicolumn{1}{l|}{0.9716} & \multicolumn{1}{l|}{0.9135}
\\ \hline
\multirow{1}{*}{STCAD-PE} & \multicolumn{1}{l|}{0.9512} & \multicolumn{1}{l|}{0.7485} & \multicolumn{1}{l|}{0.9420} & \multicolumn{1}{l|}{0.7183} & \multicolumn{1}{l|}{0.9840} & \multicolumn{1}{l|}{0.8873} & \multicolumn{1}{l|}{0.9787} & \multicolumn{1}{l|}{0.9035}
\\
\multirow{1}{*}{STCAD-SPE} & \multicolumn{1}{l|}{0.9609} & \multicolumn{1}{l|}{0.8477} & \multicolumn{1}{l|}{0.9552} & \multicolumn{1}{l|}{0.8565} & \multicolumn{1}{l|}{0.9867} & \multicolumn{1}{l|}{0.9156} & \multicolumn{1}{l|}{0.9856} & \multicolumn{1}{l|}{0.9185}
\\
\multirow{1}{*}{STCAD-TPE} & \multicolumn{1}{l|}{0.9481} & \multicolumn{1}{l|}{0.7250} & \multicolumn{1}{l|}{0.9451} & \multicolumn{1}{l|}{0.7203} & \multicolumn{1}{l|}{0.9852} & \multicolumn{1}{l|}{0.9026} & \multicolumn{1}{l|}{0.9914} & \multicolumn{1}{l|}{0.9378}
\\ \hline
\multirow{1}{*}{STCAD-SSL} & \multicolumn{1}{l|}{0.9634} & \multicolumn{1}{l|}{0.8519} & \multicolumn{1}{l|}{0.9537} & \multicolumn{1}{l|}{0.8344} & \multicolumn{1}{l|}{0.9892} & \multicolumn{1}{l|}{0.9306} & \multicolumn{1}{l|}{0.9794} & \multicolumn{1}{l|}{0.9384}
\\ \hline 
\end{tabular}
\end{table}

\subsection{Coupling Features Analysis}
\begin{figure}[!htb]
    \centering
    \includegraphics[width=13cm]{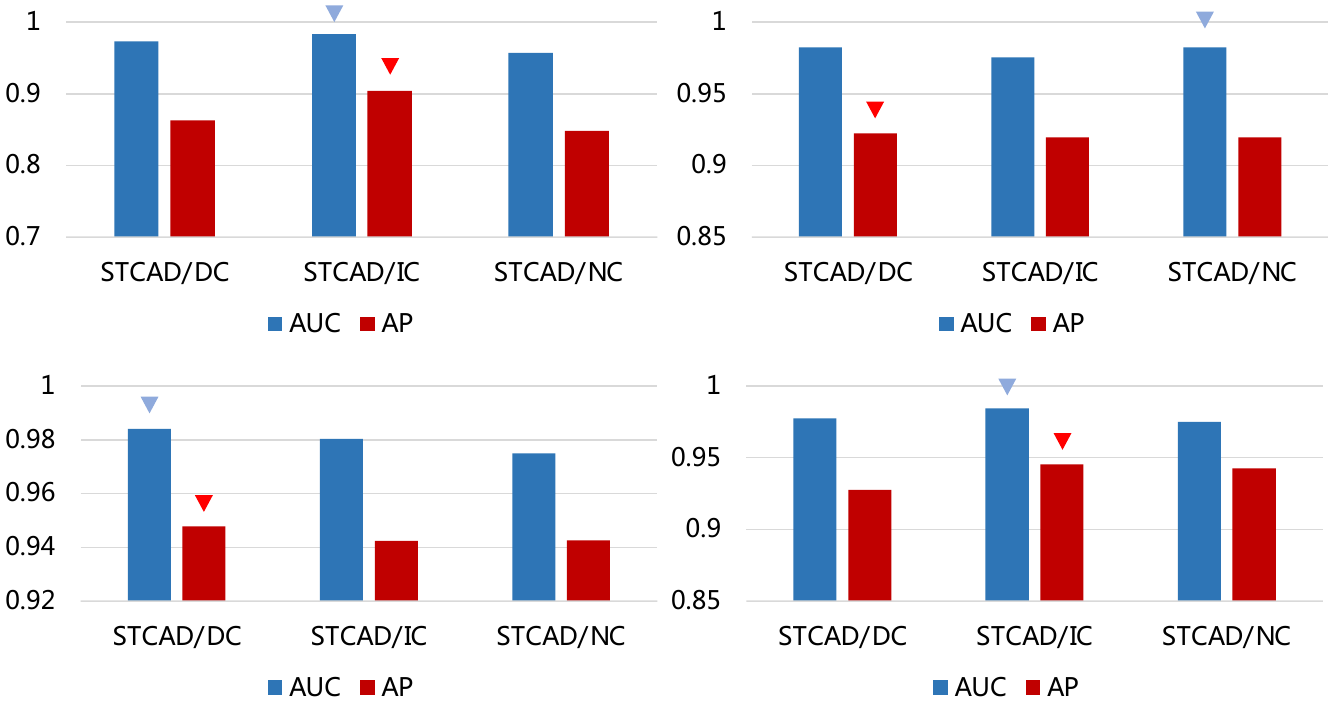}
    \caption{Impact of three coupling features on AUC and AP metrics. Four datasets are evaluated including UCI-Message (top left), Digg (top right), AS-Topology (bottom left), and Bitcoin-Alpha (bottom right). Best results on each dataset are marked with inverted triangles.}
    \label{knowledge}
\end{figure}

We next analyze structural-temporal coupling features, which capture anomaly behavior from evolutionary graphs. Three models named STCAD/DC, STCAD/IC, and STCAD/NC are implemented to only keep the feature of distance, interaction, and common-neighbor changes, respectively. The results on four datasets with the same settings are shown in Figure \ref{knowledge}. We observe that all these features are helpful in this task. However, each feature may affect diversely on different datasets. The feature that reflects interaction with surroundings is more useful in UCI-Message and Bitcoin-Alpha datasets. The distance change feature facilitates better discrimination in AS-Topology.

\subsection{Parameter Sensitivity}
\begin{figure}[!htb]
    \centering
    \includegraphics[width=13cm]{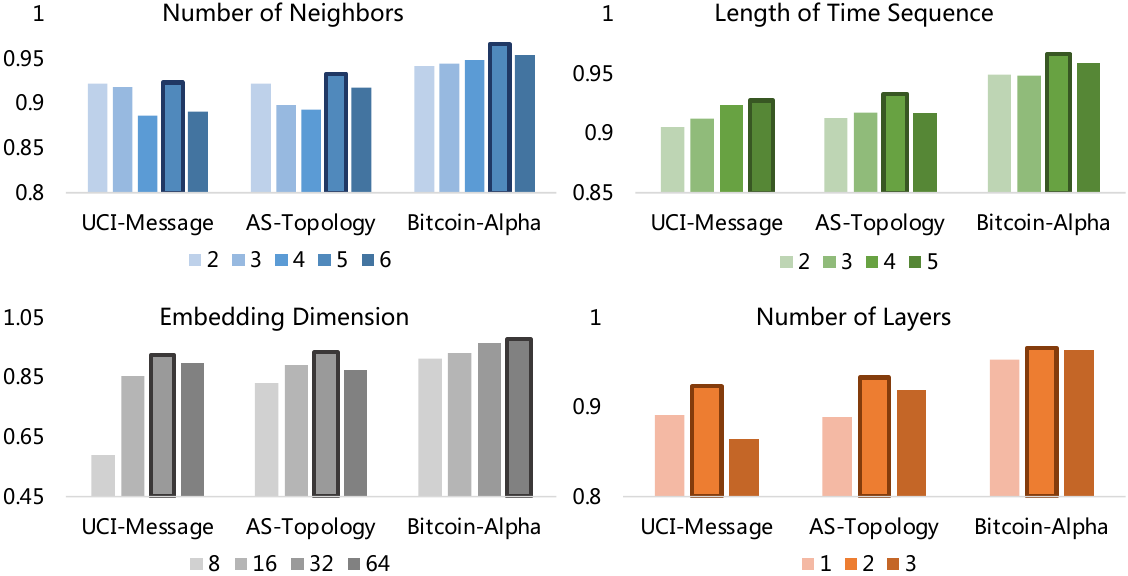}
    \caption{Performance with different hyperparameters on three datasets. The best results are highlighted with bold boxes.}
    \label{parameter}
\end{figure}

Critical hyperparameters in STCAD are investigated. We set the number of context nodes $C \in \{2,3,4,5,6\}$, the time sequence length $T \in \{2,3,4,5\}$, the embedding dimensions $D \in \{8,16,32,64\}$, and the number of attention layers $\Gamma \in \{1,2,3\}$. AP metrics on three datasets (UCI-Message, AS-Topology, and Bitcoin-Alpha) are calculated. The results are shown in Figure \ref{parameter}. We observe that better results do not always occur with more context nodes and timestamps. Setting $C=5$ and $T=4$ is a relatively better option. On the other hand, choosing appropriate transformer parameters (with $D=32$ and $\Gamma=2$) is crucial. Using too small or too large of the embedding dimensions and attention layers both degrade the performance.

\subsection{Latent Embedding Visualization}
\begin{figure}[!htb]
    \centering
    \includegraphics[width=13cm]{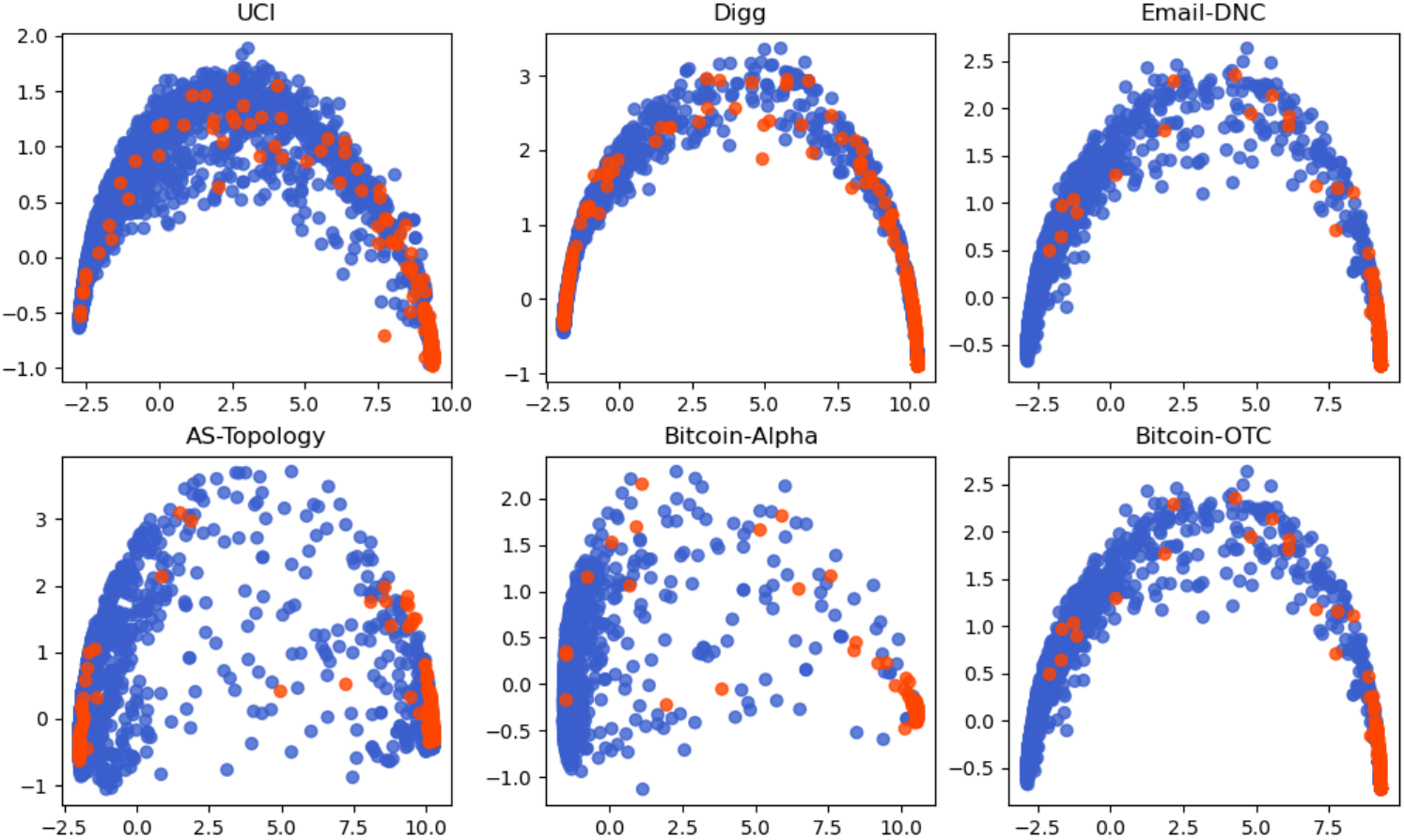}
    \caption{Latent space embeddings generated by STCAD on six datasets. Red dots are anomalous edges and blue dots are normal samples.}
    \label{embedding}
\end{figure}

To further illustrate the ability of STCAD to distinguish anomalous edges from normal ones, we visualize the embeddings of edges from the output of our transformer encoder in latent space. We train our model and record the intermediate embeddings after 200 epochs. In order to plot them in a two-dimensional space, we employ a PCA method to reduce the original dimension from 32 to 2. The visualization results shown in Figure \ref{embedding} illustrates that most of the anomalies can be easily identified by our method.

\subsection{Training Ratio Effectiveness}
\begin{figure}[!htb]
    \centering
    \includegraphics[width=13cm]{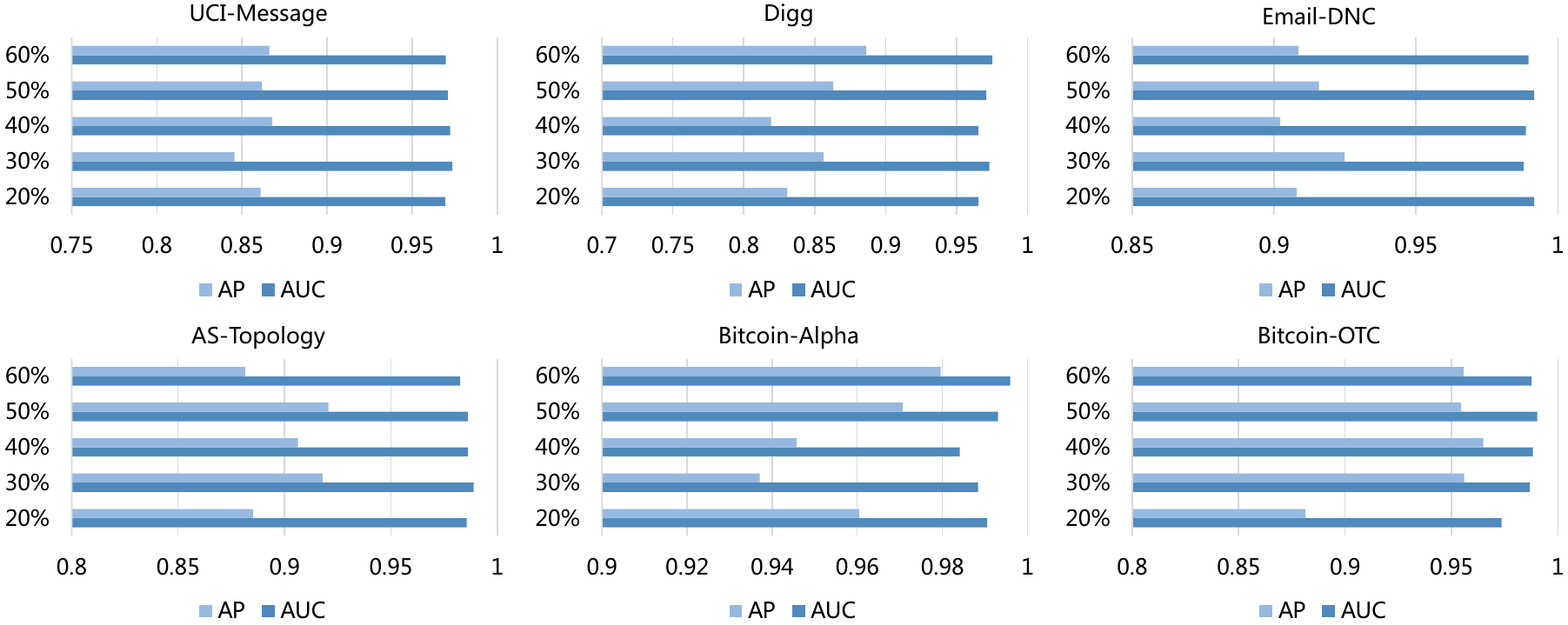}
    \caption{Effectiveness of STCAD with different training ratios on six datasets.}
    \label{ratio}
\end{figure}

In this experiment, we evaluate the effectiveness of our STCAD with different training ratios. We set the ratios to 20\%, 30\%, 40\%, 50\% and 60\% of the total training samples for each dataset. Other hyperparameters are fixed to be the same as our main experiments. The results of AUC and AP metrics are reported in Figure \ref{ratio}. We observe that our method performs competitive even with smaller ratios of training data, which demonstrates that our STCAD can efficiently utilize the coupling information from the limited number of samples. 

\section{Case Study: Emerging Technology Identification}

We apply our STCAD to a novel task to identify emerging technology fields for helping researchers find new directions. Inspired by the work \cite{krenn2022predicting} which consider finding emerging technology as a link prediction problem (new combination of existing domains), given a dynamic graph with technology fields as nodes and their co-occurrence in the same document as edges, our aim is to identify the anomalous edges so that to further discover new technology combinations.

We collect technical documents of patents and projects related to two different domains. For patents, we collect 164 AI related CPC (Cooperative Patent Classification) codes from PATENTSCOPE \footnote{https://www.wipo.int/tech\_trends/en/artificial\_intelligence/patentscope.html} and screen out patents published in the year of 2022 from a commercial website. Each CPC code is denoted as a node, and the co-occurrence of two CPC codes becomes an edge. For research projects, documents published after the year of 2020 from the National Cancer Institute are collected by using the NIH RePORTER search engine \footnote{https://reporter.nih.gov/advanced-search}. Keywords in each project are treated as nodes, co-occurrence relations are edges. We name our datasets AI-Patent and NCI-Project, respectively.

\begin{figure}[!htb]
    \centering
    \includegraphics[width=11cm]{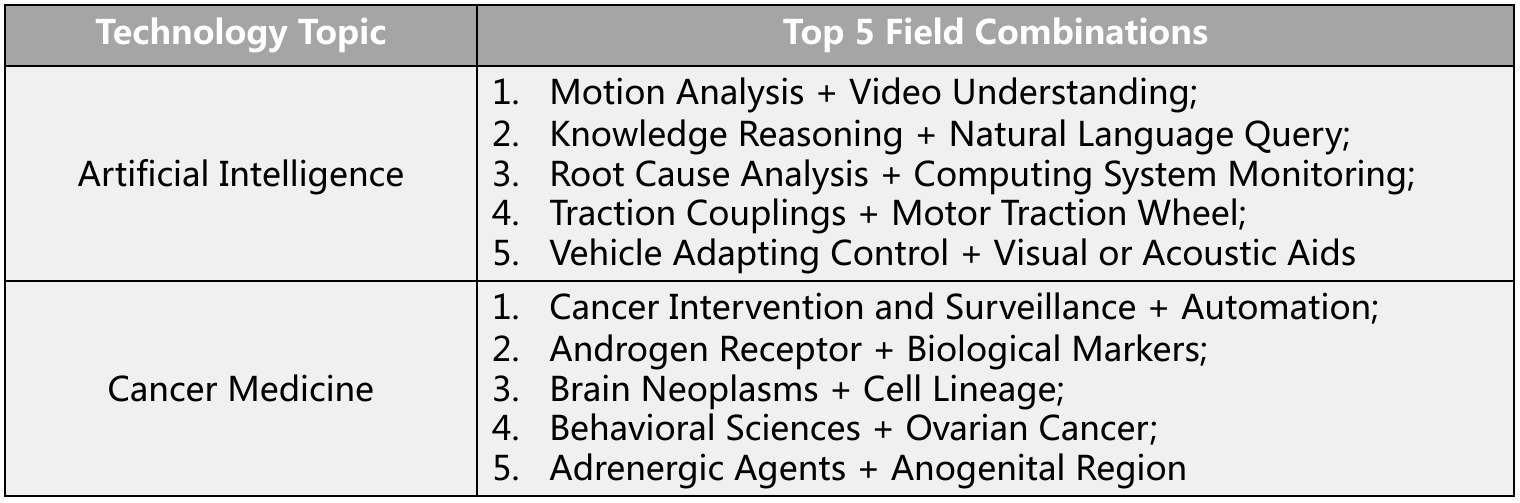}
    \caption{Top 5 combinations of technology fields in artificial intelligence and cancer medicine topics identified by our STCAD.}
    \label{technology}
\end{figure}

We export the top 5 combinations according to anomaly scores generated by STCAD, shown in Figure \ref{technology}. We manually analyze the output through searching possible evidences from the Internet. We notice that some fields are becoming popular, such as \textit{Knowledge Reasoning for Natural Language Querying} (rank 2), \textit{Visual Aided Vehicle Adapting Control} (rank 5) in artificial intelligence and \textit{Automatic Cancer Surveillance} (rank 1) in cancer medicine, which shows STCAD is useful to identify raising domains. Meanwhile, some others are completely new fields without any matching results and may lead to future research directions, such as \textit{Brain Neoplasms Cell Lineage} (rank 3) and \textit{Androgen Receptor Biological Markers} (rank 5). 

\section{Conclusion}
In this paper, we investigate the problem of anomaly detection in dynamic graphs. A structural-temporal coupling method named STCAD is proposed for detecting anomalous edges with dynamic graph transformer. It utilizes two-level anomaly-aware information from evolutionary graphs and learns integrated embeddings with a dynamic graph transformer model enhanced by 2D positional encoding. Discrimination and contextual consistency signals are captured to improve learning performance. We evaluate our method on six benchmark datasets and analyze the ability of different modules in STCAD. The results demonstrate the effectiveness of our method. A novel case study is further explored to discuss its potential ability in emerging technology identification.

\bmhead{Availability of data and code}
Source code for STCAD and baselines, and six benchmark datasets with our AI-Patent and NCI-Project datasets, are all published and available from our GitHub repository at \url{https://github.com/changzong/STCAD}.

\bmhead{Acknowledgments}
This work is supported by the Key Research and Development Program of Zhejiang Province, China (No. 2021C01013).

\section*{Declarations}
\bmhead{Conflict of interest} The authors have no conflict of interest to declare that are relevant to the content of this
article.


\bibliography{sn-bibliography}

\end{document}